\documentclass[runningheads]{llncs}

\usepackage[T1]{fontenc}
\usepackage{graphicx}
\usepackage{todonotes}
\usepackage{booktabs}
\usepackage{supertabular}
\usepackage{adjustbox}
\usepackage{float}
\usepackage{tabularx}
\usepackage{tablefootnote}
\usepackage{verbatim}
\usepackage{listings}
\usepackage{hyperref}
\usepackage{lscape}
\usepackage{csquotes}
\usepackage{amsmath}
\usepackage{xcolor}
\usepackage{cleveref}
\usepackage{subcaption}
\newcommand{\spaceunderfig}{-0.7cm}

\usepackage[inline]{enumitem}

\begin{document}
\title{Human Evaluation of Procedural \\Knowledge Graph Extraction from Text \\with Large Language Models}
%
\titlerunning{Human Evaluation of Procedural KG Extraction with LLMs}
%
\author{Valentina Anita Carriero \orcidID{0000-0003-1427-3723} \and
\\Antonia Azzini \orcidID{0000-0002-9066-1229} \and 
Ilaria Baroni \orcidID{0000-0001-5791-8427} \and 
\\Mario Scrocca \orcidID{0000-0002-8235-7331} \and 
Irene Celino \orcidID{0000-0001-9962-7193}}
\authorrunning{Valentina Anita Carriero et al.}
%
\institute{Cefriel -- Politecnico di Milano, viale Sarca 226, 20126 Milano, Italy\\
\email{name.surname@cefriel.com}}
\maketitle              
\begin{abstract}
Procedural Knowledge is the know-how expressed in the form of sequences of steps needed to perform some tasks. Procedures are usually described by means of natural language texts, such as recipes or maintenance manuals, possibly spread across different documents and systems, and their interpretation and subsequent execution is often left to the reader.
Representing such procedures in a Knowledge Graph (KG) can be the basis to build digital tools to support those users who need to apply or execute them.

In this paper, we leverage Large Language Model (LLM) capabilities and propose a prompt engineering approach to extract steps, actions, \textcolor{black}{objects, equipment and temporal information} from a textual procedure, in order to populate a  Procedural KG according to a pre-defined ontology. 
We 
evaluate the KG extraction results by means of a user study, in order to qualitatively and quantitatively assess \textcolor{black}{the perceived quality and usefulness of} 
the LLM-extracted procedural knowledge. 
\textcolor{black}{We show that LLMs can produce outputs of acceptable quality and we assess the subjective perception of AI by human evaluators.} 





\end{abstract}

\section{Introduction}
\label{sec:intro}
Procedural Knowledge (PK) is knowing how to perform some tasks. 
Such knowledge is usually expressed in the form of sequences of steps needed to achieve an overall goal, as in the case of recipes and maintenance activities.
Making PK explicit is not trivial: identifying all actions to be performed, and splitting them into separate steps including all relevant information (e.g., the needed equipment), is often subject to interpretation and commonsense.
Moreover, even when documented, this is usually done in an unstructured format, by means of natural language, across heterogeneous documents and systems. 
This makes it hard to access procedures for those who need to execute them, leading to partial or poor compliance with processes and best practices.
As argued in \cite{qurator22}, formally representing such procedures as Knowledge Graphs (KGs) can be the basis to build digital tools that provide automatic support to the users, and foster procedural knowledge documentation, reuse, and interoperability.

However, developing high-quality KGs is time-consuming: making this task even partially automatic may be a game changer.
Large Language Models (LLMs) -- probabilistic models trained on huge natural language corpora -- exhibit remarkable capabilities in natural language processing and generation.
LLMs have recently gained significant attention to support different tasks of the whole knowledge engineering process, as emphasized in the report of the Dagstuhl Seminar 22372 \cite{groth_knowledge_2023} and the LM-KBC Challenge\footnote{Cf. \url{https://lm-kbc.github.io/}}.

In this paper, we leverage LLM capabilities for extracting PK from text; differently from other extraction tasks like entity recognition or relation extraction, identifying procedural steps (the actual sentences describing actions to be performed) is a potentially more ambiguous task: as we will show in our experiments, different human annotators may produce different yet valid outcomes, thus precluding the existence of a definitive ground truth for such task. 

%

Our contribution is therefore twofold. On the one hand, we experiment different prompting strategies to employ LLMs for extracting PK from unformatted textual descriptions, and generating a valid KG thereof according to a predefined procedural ontology. On the other hand, we evaluate the knowledge extraction outcomes by means of a user study, in order to evaluate the (perceived) quality and  usefulness of the resulting KG, comparing it to human annotation and assessing the user-acceptance of the LLM-produced results on such a subjective task\textcolor{black}{; we also evaluate a potential bias in human evaluation with respect to AI}.
%
All materials, code, results of this work are available on GitHub\footnote{Cf. \url{https://github.com/cefriel/procedural-kg-llm}}.

The paper is organised as follows: after introducing the state of the art in Section~\ref{sec:related}, we better define the problem and the data we used in Section~\ref{sec:input}; then we report the findings of a preliminary study in Section~\ref{sec:preliminary} which led us to the definition of our research questions in Section~\ref{sec:RQs}; Section~\ref{sec:meth-rel} describes our iterative prompt engineering process, and the results of our proposed prompt-based pipeline; in Section~\ref{sec:setting} we describe our human evaluation design and we present and discuss our quantitative and qualitative findings in Section~\ref{sec:eval}; Section~\ref{sec:concl} concludes the paper with some future work.

\section{Related Work}
\label{sec:related}

Our work is at the intersection of the LLM application to Knowledge Engineering and its result assessment from a human point of view. Therefore, we take into account several aspects from the state of the art, as explained in the following.

\subsection{LLMs for Knowledge Engineering and Knowledge Extraction}
As reported in \cite{allen_knowledge_2023}, LLMs provide a powerful tool for mapping natural language to formal language, a fundamental activity in knowledge engineering.
Existing efforts in unifying LLMs and KGs are summarized in~\cite{pan_2024,pan_large_2023}. The LM-KBC challenge explores how to build disambiguated knowledge bases from LLMs, given a subject and a relation (see, among others, \cite{yang_2023,LLMKE_2023}).

Several recent works cover the use of LLM prompting to address specific knowledge engineering tasks, like ontology engineering~\cite{zhang2024ontochat,ciroku2024revont,bischof2024ontoterms,hoseini2024towards}, ontology learning~\cite{fathallah2024neongpt,babaei2023llms4ol}, named entity recognition and linking~\cite{kumar2020,ding2021few,shi2024generative}, knowledge graph construction including mapping generation~\cite{li2023knowledge,hofer2024towards}; some works specifically focus on benchmark, metrics and evaluation of such methods~\cite{text2kgbench2023,frey_2023}.

Specifically on (procedural) knowledge extraction, several studies (including \cite{zhou_learning_2019,zhang_reasoning_2020,zhang_reasoning_2022}) investigate some extraction and reasoning tasks on procedures, such as the relation between a step and the procedure's goal, and the temporal relation between steps. 
Other works focus on the business process management domain: a corpus of business processes annotated by humans with activities, gateways, actors, and flow information~\cite{bellan2021process} is exploited in~\cite{bellan2022extracting} for extracting, from such processes, activities and their participants, using an LLM; an activity recommender to support business process modeling is introduced in~\cite{sola2023activity}. 
A semi-structured dataset of repair manuals related to \textit{Mac Laptops} is annotated with the required tools and the parts that are disassembled during the process, along with two methods for extracting them~\cite{nabizadeh2020myfixit}. The procedures included in the dataset focus on opening the device and removing/repairing a broken component.
Reusing such datasets as-is was not applicable for our experiments, since they did not fully covered our requirements, e.g., by taking into account only actions related to repairs~\cite{nabizadeh2020myfixit}, and by missing annotations about objects and tools~\cite{bellan2021process} (cf. Section \ref{sec:input}).
Micrographs storing relevant entities and actions from technical support web pages are created in~\cite{kumar2021}. 
A prompting-based pipeline to extract a list of ordered steps from a procedure expressed as a numbered/bullet/indented list is proposed in~\cite{rula_procedural_2023}; \textcolor{black}{differently, we give as input a text with no formatting and include additional descriptive sentences that the LLM needs to discard}.

\subsection{Evaluation of Generative AI}
The evaluation of Generative AI and specifically LLMs is an open research problem, in that it is not easy to objectively assess the generated (natural language) output~\cite{liu2023g}. In the area of text annotation, different approaches exist~\cite{brown2020language,su2022selective}, because LLMs are expected to act like human annotators~\cite{marreddy24krtutorial}. Indeed, among other things, they can be employed for fact checking~\cite{ni2024afacta} or to identify claims and sub-claims in text entailment~\cite{kamoi2023wice}.

The approach to assess LLM outputs can combine human and LLM annotations~\cite{li2023coannotating}, and several authors propose to use LLMs to perform also the evaluation task: by simulating human feedback via LLMs~\cite{dubois2024alpacafarm}, by employing a (LLM) debate method~\cite{chan2023chateval}, or by making LLMs adopt a human-like comparative evaluation approach~\cite{yuan2023batcheval}.

While a unified approach to the assessment of natural language generation has not yet emerged, some work started to propose evaluation frameworks~\cite{fu2023gptscore}, also focusing on the definition of evaluation metrics for specific tasks like summarization~\cite{liu2022revisiting}. \textcolor{black}{We based our human evaluation approach on these works, as well as on general usability methods.} It is important to note that a holistic evaluation should not only assess the quality of the LLM outcome, but also the human interaction experience with the LLM~\cite{lee2022evaluating}.

\subsection{Human Bias on (Generative) AI}
Finally, whenever involving humans in the evaluation of an artificial system, the subjective perception and the potential cognitive biases of users must be carefully considered. Indeed, some studies suggest that human attitudes towards AI are largely negative~\cite{neudert2020attitudes}. In the area of Generative AI, several concerns emerge in relation to the creativity potential of such models~\cite{magni2023humans}, especially in the context of art~\cite{ragot2020ai,millet2023defending} and music~\cite{zlatkov2023searching}.

\textcolor{black}{In our work, we also investigate the potential bias of human subjects against LLMs that play the role of annotators, by applying an A/B testing approach,} as done by~\cite{hidalgo2021humans}: they performed several experiments on how people judge humans and machines differently, in several scenarios (e.g., natural disasters, labor displacement, policing, privacy, algorithmic bias); they demonstrate that people tend to favour humans or machines in different scenarios, revealing the biases in human-machine interaction.

\section{Problem Definition and Data}
\label{sec:input}

As mentioned in the introduction, our goal is to build a knowledge graph out of textual descriptions of procedures. 
Our vision is therefore to extract procedural knowledge (PK) from those documents, and to build a knowledge graph (KG) according to an ontology. This KG can then be used by different downstream applications to facilitate the access and use of such procedures by human operators. Examples of such applications could be (KG-empowered) search applications or intelligent assistant: we expect the users to feel the need to be helped to find a specific procedure (or a part thereof) and to be guided step-by-step in its execution, for example by being informed about the action they have to perform (e.g., \emph{turning off} a switch), the equipment they may need to use (e.g., wearing \emph{protective gloves}) or the time it may take to perform a specific step (e.g., approximately \emph{15 minutes}). 

In order to fulfill such requirements, the procedural KG extracted from text should (1) preserve the intended meaning of the original document and (2) contain enough information to guide a user in correctly executing the procedure. The extracted KG should therefore be evaluated, respectively, on the basis of its \emph{quality} and \emph{usefulness} (cf. Section~\ref{sec:setting}).
Based on this scenario, we define a simple ontology and we identify a general-purpose dataset to be used in our LLM-powered PK extraction and KG building experiments.

\begin{figure}[tb]
    \centering
    \caption{Ontology used in the experiments.}
    \includegraphics[width=.8\linewidth]{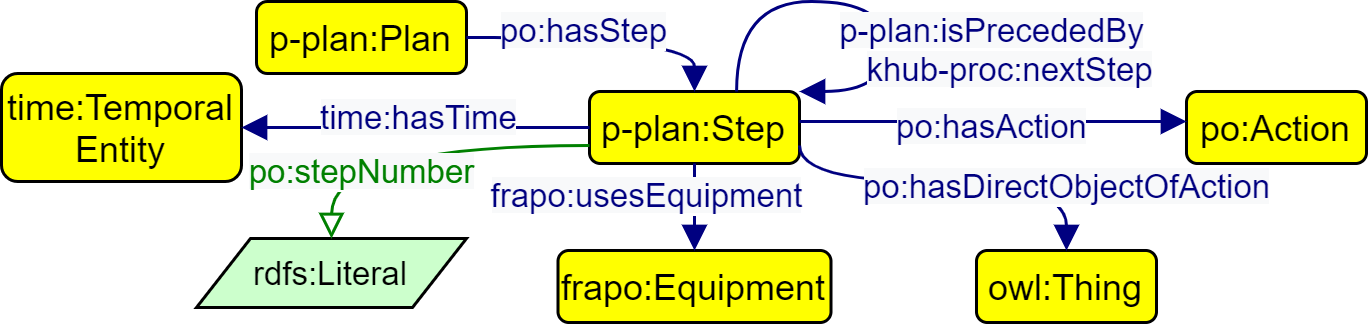}
    \label{fig:ontology}
\vspace{\spaceunderfig}
\end{figure}

\textbf{Ontology.} We reuse existing ontologies when applicable, namely: P-Plan \cite{garijo2012augmenting} and K-Hub~\cite{RulaCABC23}, that address plans and related concepts, FRAPO, one of the SPAR ontologies\footnote{Cf. \url{http://www.sparontologies.net/}}, and the Time Ontology\footnote{\texttt{p-plan:} http://purl.org/net/p-plan\# \\ \texttt{khub-proc:} https://knowledge.c-innovationhub.com/k-hub/procedure\# \\ \texttt{frapo:} http://purl.org/cerif/frapo/ \\ \texttt{time:} http://www.w3.org/2006/time\#}, while creating a few new classes and properties when needed (\texttt{po:}). 
Specifically, as depicted in Figure \ref{fig:ontology}, a procedure is represented by the class \texttt{p-plan:Plan} and is linked to its \texttt{p-plan:Step}s (\texttt{po:hasStep}), which are sequentially ordered (property \texttt{p-plan:precededBy} and its inverse \texttt{khub-proc:nextStep}). Each step is then linked to its \texttt{frapo:E\-quip\-ment}, if any, via the property \texttt{frapo:usesEquipment}, and with the action(s) to be performed while executing it (\texttt{po:hasAction}), along with the direct object of the action (\texttt{po:hasDirectObjectOfAction}). The information about the time needed for executing the step is represented by the class \texttt{time:TemporalEntity}.

\textbf{Input procedures.} For our experiments, we use as a reference dataset Wiki\-How,
one of the largest online databases of PK, which includes how-to articles on multiple domains. 
We reuse the JSON dataset built by \cite{zhang_reasoning_2020}, crawled from the WikiHow website\footnote{Cf. \url{https://github.com/zharry29/wikihow-goal-step}}, and focus on atomic procedures, that do not include methods/parts as sub-procedures.
Out of each selected procedure, we build an unformatted text (also partially removing punctuation), by concatenating (i) procedure title, representing the overall goal, (ii) general procedure description, (iii) headline of each step, (iv) description of each step.
Thus, we also include in the text irrelevant sentences, which are supposed to be discarded during the extraction phase.
In total, we used four procedures randomly selected from WikiHow, one as working example for our prompt engineering phase, and three for our method replication \textcolor{black}{and human assessment}. Such procedures are diverse with respect to both complexity and topic: 
\emph{how to clean a computer monitor} (working example), \emph{how to fix a rubbing door}, \emph{how to make honey glazed parsnips}, and \emph{how to plant a bare root tree}.



%
%
\section{Preliminary Study}
\label{sec:preliminary}
In a preliminary formative study, we wanted to kick-start the LLM-powered procedure extraction from text, by identifying the best performing prompt chaining~\cite{wei2022-cot-reasoning} to execute the task. We evaluated both the prompt instructions and the LLM outputs ourselves, but we also wanted to involve human evaluators outside our research team to get an external assessment. 
\textcolor{black}{Employing the same procedural texts described in Section~\ref{sec:input} and a  prompting approach similar to the one described in~\ref{sec:meth-rel},} we extracted the procedural knowledge from text according to the predefined ontology; the procedure annotation task was explained (both to the human evaluators and the LLM) as a verbatim extraction from the text of the relevant sentences explaining the procedure steps and the identification of specific parts of those steps (e.g., action, tools). Then, we involved a group of people to judge the annotation of each procedure (n=24, 8 evaluators for each procedure; \textcolor{black}{79\% male, 21\% female; aged between 20 and 60}); each participant was asked to first manually perform the entire extraction (generating a semi-structured representation, not the final RDF format) and then to evaluate the LLM results on the same procedure and to give us suggestions on how to improve the process. 
The results of this study were partially unexpected and were very useful to design the experiment described in this paper. Our preliminary observations were as follows.
\textbf{Humans highly disagree in annotating procedural knowledge.} The number of annotated steps, tools, and actions varied quite a lot, thus the level of agreement of people in executing such a task was very low (no significant inter-rater agreement indicator). 
This suggests that \emph{not all annotation/extraction tasks are equal}: in our case, 
the step identification is quite different from an entity recognition or relation extraction task, because it involves the extraction of entire sentences, thus our goal is somehow more \enquote{subjective} and it is not amenable to have an actual \enquote{ground truth}. This also means that \emph{the LLM can be considered as \enquote{yet another annotator}}.

\textbf{Humans tend to adjust the task regardless of the instructions and example.} Analysing their annotations, we realised that humans did not closely followed the given instructions (opposed to the LLM \enquote{obedient} execution) based on what they consider more useful, using their creativity or reasoning. In some cases, new steps that were absent from the input procedure were introduced; some actions were inferred by the participants; some users tended to rephrase or summarise the text of the step, to make it clearer based on their interpretation, even if we asked for verbatim extraction of sentences.
Moreover, some participants were quite critical with respect to the prompt instructions, commenting that they were too long and not immediate.
This suggests that the prompt can be less detailed and \emph{leaves room for the more generative capabilities of the LLM}, as humans do the same.

\textbf{Humans are highly critical when evaluating the LLM.} 
Also in the evaluation of LLM results, performed through a list of Likert-scale ratings on various aspects of the generated output, there was no agreement between the participants (computed as Krippendorff's alpha metrics~\cite{hayes2007answering}), as they directly compared their work with that of the LLM which was quite different. Some evaluators 
assigned very low scores, which means that the LLM is still not perceived by some as \enquote{good enough} to solve the task. 
This suggests that those \emph{evaluations may have been influenced by a bias on AI} and that this hypothesis was worth being tested.
\section{Research Questions}
\label{sec:RQs}
Our goal is to create a procedural knowledge graph by extracting structured knowledge from unstructured text; 
we would like to apply LLMs to the \emph{procedural knowledge extraction} task and assess their results from a human point of view, taking into consideration the findings of our preliminary study.

Our research questions are formulated as follows:
\begin{itemize}
    \item[RQ1] \emph{What is the quality of the extracted procedural knowledge (graph) as perceived by human evaluators?} In this respect, we would like to understand if people judge the LLM-extracted knowledge as correctly representing the meaning of the original text and what influences their evaluation.
    \item[RQ2] \emph{What is the usefulness of the extracted procedural knowledge (graph) as perceived by human evaluators?} In this respect, we would like to assess the  \enquote{fitness for use} of the \textcolor{black}{output of the LLM extraction}, 
    when people are explained its potential downstream use.
    \item[RQ3] \emph{Do human evaluators show any systematic bias if they are told that the extraction task was executed by a LLM rather than an expert human annotator?} In this respect, we would like to check any difference in people judgment about the extracted procedural knowledge (graph).
    \item[RQ4] \emph{Qualitatively, are there any differences in the way human annotators and the LLM extract explicit knowledge and infer implicit knowledge from the text?} In this respect, we would like to assess if the LLM behaves like human annotators in interpreting and extracting procedural knowledge. 
\end{itemize}

%
\section{Prompt Engineering Solution}
\label{sec:meth-rel}
The expected result for each procedure was an ordered list of sentences summarizing the steps extracted from the text, with their respective actions, direct objects, equipment items, and temporal information, expressed in RDF according to the given ontology. We tested different prompting approaches, as detailed below, and in an initial phase we manually evaluated the LLM results, finding an agreement between us with majority voting to decide whether each result was qualitatively satisfactory; the prompting approach that revealed to yield the best results was then used in the subsequent human evaluation experiment (cf. Section~\ref{sec:setting}).

\noindent \textbf{Prompting approach.}
We tested different prompting approaches on the working procedure (cf. Section \ref{sec:input}) to find the best solution. Initially, we tried with a single prompt with zero/one/few-shot and we found that one prompt was not enough to achieve an acceptable result and a correct RDF, while giving one example in the prompt (one-shot learning) was the best trade-off between zero-shot and few-shots; then, as demonstrated in \cite{wei2022-cot-reasoning}, we tried a Chain-of-Thought (CoT) prompting approach, i.e. decomposing the global problem into intermediate steps (prompts) and solve each of them sequentially, because this significantly improves the ability of LLMs to perform complex reasoning tasks, including commonsense.
We tested our prompts with or without the definitions of the entities to be extracted, and found that very short definitions 
worked better in obtaining a more accurate result than both longer definitions 
and no definition at all.
Moreover, allowing the LLM to generate the step sentence descriptions gave better results in the extraction of steps as opposed to strictly report the verbatim sentence from the text without modifications, as in the preliminary experiment (cf. Section~\ref{sec:preliminary}).
After multiple refinement iterations,
we came up with a CoT prompting approach decomposed in 2 steps, both applying a one-shot learning. 
Each step is associated with a different agent playing a different role.

\begin{figure}[tb]
    \centering
    \caption{Example procedure and output provided in prompt \emph{P1}.}
    \includegraphics[width=.8\linewidth]{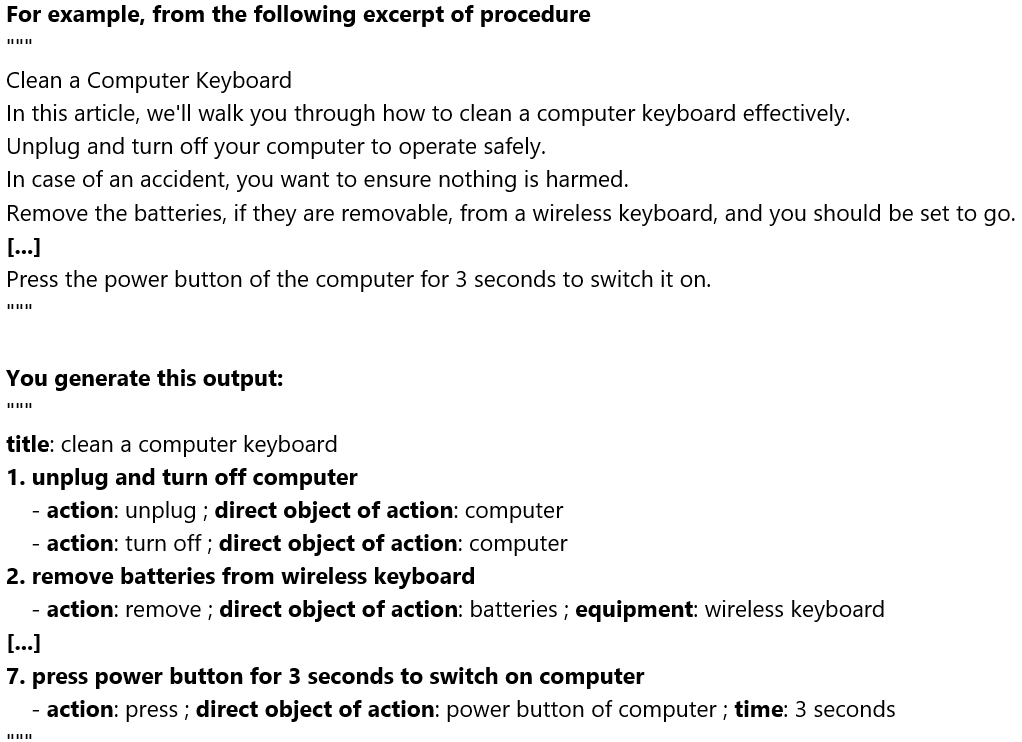}
    \label{fig:example-proc}
\vspace{\spaceunderfig}
\end{figure}

\emph{P1: generate steps descriptions and annotations.} In this prompt, the LLM is given the role of \enquote{expert in information extraction with a special background in procedures} and is asked to generate a semi-structured output, composed of a list of sentences describing the procedure steps, each including the performed action(s), its direct object(s), the used equipment, and any execution time information (if any).
Each piece of information to be included in the semi-structured output is explained as follows: 
\begin{itemize}
\item an \emph{action} is defined as \enquote{the verb or phrasal verb or idiomatic expression that represents an action to be performed},
\item the \emph{direct object} of the action is defined as \enquote{the noun or pronoun being acted upon by the verb of the action},
\item the \emph{equipment} is an \enquote{item that is needed to perform the action},
\item the \emph{time} is \enquote{any temporal information relevant to perform the action}.
\end{itemize}
It is further specified in the prompt that each generated step may include one or more actions, and that the direct object of each action cannot be considered an equipment itself.
Finally, we asked the LLM to include also possibly implicit equipment that is not mentioned in the text, if relevant to perform the action.
By asking the LLM to generate a list of steps from an input procedure, that summarizes the input and contains a specific subset of information, we managed to leverage its generative capabilities. 
The prompt also includes an example of procedure with its respective output (see Figure \ref{fig:example-proc}). The example is crucial to show the LLM not only how to generate the semi-structured output, but also how to produce the annotations that are preliminary to the RDF KG construction. 

\emph{P2: generate the ontology-based knowledge graph of the procedure.} In order to obtain the intended output -- i.e. a KG of the procedure, linked to its steps, actions, direct objects, equipment and temporal information, according to the given ontology -- we assign to the LLM the new role of \enquote{expert in knowledge graph construction, with a special background in ontologies on procedural knowledge}, and we ask it convert the semi-structured output of the first prompt into RDF formatted in Turtle syntax (similarly to \cite{frey_2023}).
Rather than providing the entire ontology to be used, we showed the language model an example translation from its initial output to RDF, so that the LLM could find in the example all classes and properties to be used, and how they needed to be mapped to the annotation.

While designing \emph{P2}, we started by providing the whole ontology, asking the LLM to translate its initial output to Turtle according to such ontology.
However, relevant triples were missing: many inverse relations included in the ontology were never used, and the procedure was linked only to the first step. Moreover, additional and very specific instructions would have been needed in the prompt for defining all prefixes and how to design local names, to prevent ambiguous individuals.
Instead, by showing an example on how to map the textual list of steps and annotations to Turtle, we avoid possible errors in complying with ontological constraints, with a greater guarantee of the final KG to be complete.
Moreover, this method may be particularly useful when we work with multiple, and possibly large, ontologies, that would exceed the context window of the LLM.

\textbf{Results.}
For our experiments, we use the GPT 4o model, and rely on the LangChain framework\footnote{Cf. \url{https://www.langchain.com/}}.
We also tested with GPT 3.5 Turbo model, however, we obtained worse results than GPT 4o, with respect to both \emph{P1} and \emph{P2}. 
Specifically, GPT 3.5 Turbo tended to collapse some objects and equipment items instead of keeping them separate as in the example (e.g. \emph{bucket filled with water} in place of \emph{bucket} and \emph{water}), and it did not always comply with the provided template, as opposed to GPT 4o.
The results of both prompts are available on GitHub.
All produced KGs are valid RDF in Turtle syntax. 
Properties' domains and ranges are correctly used, and individuals are typed flawlessly. The LLM perfectly mimics our conventions for local names.
The KGs generated with \emph{P2} are complete, meaning that they include all entities that have been extracted in \emph{P1}, thus addressing all requirements we defined. 
Therefore, we use the results of \emph{P1} in our user experiment.





\begin{figure}[tb]
    \centering
    \caption{Illustration of the Chain of Prompt for Procedural KG extraction, along with an example of extracted step.}
    \includegraphics[width=\linewidth]{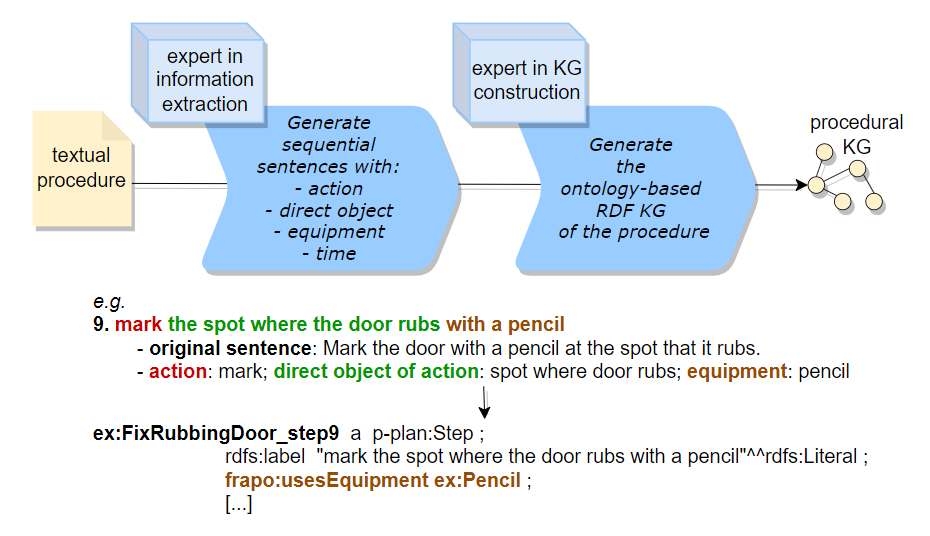}
    \label{fig:CoT}
\vspace{\spaceunderfig}
\end{figure}
\section{Experiment Design}
\label{sec:setting}
In order to test our hypotheses, we extracted PK from 3 procedure texts applying the prompting approach described in Section~\ref{sec:meth-rel}. Then, we setup a crowdsourcing campaign to collect feedback from human evaluators as follows.

\textbf{Input to the human evaluator.} Each participant was given the annotation instructions (corresponding to the prompt \emph{P1}, including an example of the expected input-output), the procedure text and the extracted semi-structured PK as generated by the LLM.

\textbf{Expected output.} Each participant was presented with a form (implemented with Microsoft Forms) and asked to perform two tasks (presented in two following pages in the form): (1) a short manual annotation exercise, to get familiar with the task and to provide feedback on the task itself, and (2) the assessment of the LLM annotation, according to the evaluation criteria explained in the following.

\textbf{Instructions.} For the manual annotation task, participants were given exactly the same instructions included in the \emph{P1} prompt used with the LLM; it is worth noting that, as explained in Section~\ref{sec:meth-rel}, the second prompt \emph{P2} always yielded a correct RDF representation of the semi-structured output of the first prompt, therefore we gave the participants the LLM results in response to \emph{P1}, to avoid the need for people to understand and validate Turtle. 
For the LLM assessment, participants were asked to express their evaluation with a set of Likert-scale ratings according to our evaluation dimensions. 

\textbf{Evaluation dimensions.} All quantitative evaluation items were expressed as statements to be assessed on a 1 to 5 Likert scale of agreement (from strongly disagree to strongly agree); we also asked for free-text feedback to also collect some more qualitative feedback. Regarding the manual annotation task performed by the participants, we asked feedback about the task difficulty and the instructions clarity, as per the items in Table~\ref{tab:ann-eval}. 
\begin{table}[tb]
\centering
\caption{Evaluation items used to assess the manual annotation task.}
\begin{tabular}{|l|}
\hline
I found this task very easy.                                  \\ \hline
I am confident that my output is correct.                           \\ \hline
I found the instructions provided for this annotation task very clear. \\ \hline 
\end{tabular}
\vspace{.1cm}

\label{tab:ann-eval}
\end{table}
Regarding the LLM output assessment, we collected ratings about three main dimensions: \emph{perceived quality}, intended as the perceived value of the knowledge extraction, \emph{perceived comparative quality}, intended as the value compared to what the evaluator would have done on the same task, and \emph{perceived usefulness}, intended as the characteristics of the output to be used as intended; the respective items are listed in Table~\ref{tab:llm-eval}. Additionally, we also asked some profiling question, to collect user characteristics, to check their possible influence on the answers. 
\begin{table}[htb]
\centering
\caption{Evaluation items used to assess the LLM output. Items marked with * are expressed on a reversed scale.}
\begin{tabular}{|p{3.1cm}|p{8.8cm}|}
\hline
\textbf{Dimension}          & \textbf{Item}                                                                                                                                            \\ \hline
Quality (Q1)            & The annotator was   accurate in rephrasing and extracting the procedure from the text                                                           \\ \hline
Quality (Q2)             & The steps identified and   rephrased by the annotator are relevant to the task to be executed                                                   \\ \hline
Quality (Q3)             & The steps identified and rephrased by the   annotator provide sufficient details to execute the procedure                                       \\ \hline
Comparative Quality  (CQ1) & I would have identified a   different set of equipments from those suggested by the annotator *                                                 \\ \hline
Comparative Quality (CQ2) & I would have identified a   different set of actions from those suggested by the annotator *                                                    \\ \hline
Comparative Quality (CQ3) & I would have identified a   different set of direct objects of action from those suggested by the   annotator *                                 \\ \hline
Comparative Quality (CQ4) & I think that the annotator   missed some important parts of the procedure in their processing *                                                 \\ \hline
Usefulness  (U1)        & Following only the instructions   identified by the annotator would make it easier to execute the procedure   (with respect to the original text)           \\ \hline
Usefulness  (U2)        & Following only the instructions   identified by the annotator would make it quicker to execute the procedure   (with respect to the original text)          \\ \hline
Usefulness   (U3)       & I believe that an automatic system   providing information for executing a procedure could use the extraction from   this annotator as an input \\ \hline
Usefulness  (U4)        & The instructions identified by the   annotator don't include the proper information to support procedure execution *                            \\ \hline
Usefulness (U5)         & I think I would correctly execute the   original procedure based only on the instructions identified by the annotator                           \\ \hline
\end{tabular}
\label{tab:llm-eval}
\end{table}
%

\textbf{Experimental setting.} For each of the procedures used in the experiment, we set up two crowdsourcing campaigns on the Prolific platform\footnote{Cf. \url{https://www.prolific.com/}}~\cite{palan2018}, in a A/B testing with a between-subject design: in one campaign, we told the participants that the annotations they were evaluating were performed by a LLM, while the other group was said that the annotations were produced by an experienced human annotator. Each campaign involved 30 respondents, accounting for a total of n=180 participants in 6 campaigns. The inclusion criteria were: age between 20 and 60, residence in Europe, fluency in English and the use of a desktop for the study execution, while, to implement an A/B testing, the exclusion criterion was the participation to the other five experimental campaigns. No inclusion or exclusion criteria were imposed on other personal characteristics (gender, nationality, etc.).

\section{Human Assessment and Discussion}
\label{sec:eval}
In this section, we summarise and discuss the results and main findings of our quantitative and qualitative evaluation with the crowdsourcing participants.

\textbf{Human Evaluators.} As explained before, we involved 180 participants from the Prolific crowdsourcing platform, between June and July 2024. Excluding participants who revoked the consent to register demographic data, the others were aged between 20 and 60, 37\% male, 63\% female; their employment status was 14\% student, 47\% employed and 39\% unemployed. There was no significant difference between the 6 groups that were involved in the 6 campaigns, in terms of demographic and profiling characteristics.

\begin{figure}[htb]
    \centering
    \caption{\textcolor{black}{Distribution of human ratings on the evaluation items (cf. Table~\ref{tab:llm-eval}).}}
    \includegraphics[width=1\linewidth]{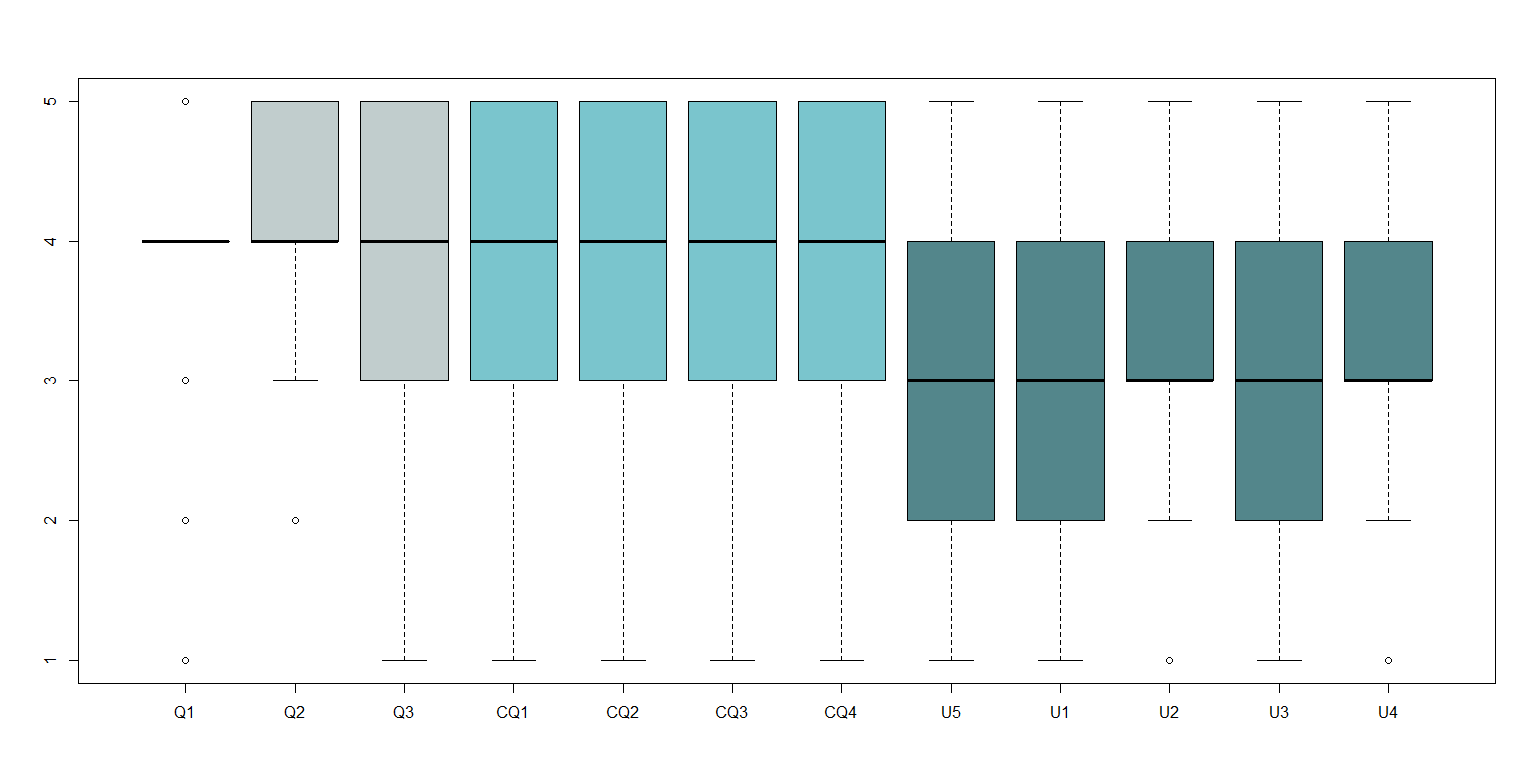}
    \label{fig:boxplot}
    \vspace{\spaceunderfig}
\end{figure}

\begin{table}[htb]
\centering
\caption{Distribution of evaluation scores by the human participants (n=180) on the items (cf. Table~\ref{tab:llm-eval}).}
\begin{tabular}{|c|c|c|c|}
\hline
Evaluation Item & Average & Median & Std dev \\ \hline
Q1              & 3,94    & 4      & 0,88    \\ \hline
Q2              & 4,16    & 4      & 0,79    \\ \hline
Q3              & 3,82    & 4      & 0,97    \\ \hline
CQ1             & 3,83    & 4      & 1,15    \\ \hline
CQ2             & 3,71    & 4      & 1,18    \\ \hline
CQ3             & 4,02    & 4      & 1,09    \\ \hline
CQ4             & 3,55    & 4      & 1,29    \\ \hline
U1              & 2,90    & 3      & 1,09    \\ \hline
U2              & 3,04    & 3      & 1,17    \\ \hline
U3              & 3,35    & 3      & 1,06    \\ \hline
U4              & 3,39    & 3      & 1,18    \\ \hline
U5              & 3,31    & 3      & 1,13    \\ \hline
\end{tabular}
\label{tab:eval-scores}
\end{table}

\textbf{Perceived Quality.} The overall distribution of the ratings given by the human evaluators on the quality-related dimensions are displayed in the first 7 boxplots in Figure~\ref{fig:boxplot} and the first 7 rows in Table~\ref{tab:eval-scores}. On the 1-5 Likert scale, the human judgement is generally positive (median value is always 4). We tested with ANOVA the possible variability of the ratings with respect to the 3 different procedures, but we did not find any strongly significant difference, which means that the results are not dependent on the different input.
With respect to \emph{perceived quality}, i.e. assessment of the extracted knowledge accuracy and relevance, we notice that the evaluators gave generally high scores 
and those scores are quite positively correlated (\textcolor{black}{$r$ between 0.56 and 0.64}). 
With respect to \emph{perceived comparative quality}, i.e. assessment of the extracted knowledge quality with respect to what the evaluator would have done on the same task, the participants were still positive, but slightly less generous in their ratings 
and those scores are a bit less positively correlated (\textcolor{black}{$r$ between 0.26 and 0.59}); this means that the participants still think that the quality of the extracted knowledge is quite high, but some of them suppose that they could have done a better job, confirming the findings of our preliminary study (cf. Section~\ref{sec:preliminary}). 

We can therefore conclude that, in relation to our research question \textbf{RQ1}, on average, human evaluators judge the extracted knowledge as correctly representing the meaning of the original text.

\textbf{Perceived Usefulness.} The overall distribution of the ratings given by the human evaluators on the usefulness items are displayed in the last 5 boxplots in Figure~\ref{fig:boxplot} and the last 5 rows in Table~\ref{tab:eval-scores}. On the 1-5 Likert scale, the human judgement ratings are generally lower that the quality-related items (median value is always 3)  and the scores are weakly or moderately correlated (\textcolor{black}{$r$ between 0.15 and 0.62}).  Those results indicate that the evaluators were more doubtful when they had to judge the usefulness; this may be due to an unclear explanation or an imprecise understanding of the intended use of the procedural knowledge extracted from the text. Indeed, 
when asked to assess their manual annotation and the clarity of the instructions they gave central-range scores (average values between 2.84 and 3.19 on a 1-5 scale, cf. items in Table~\ref{tab:ann-eval}): this may have negatively impacted their interpretation of the \enquote{fitness for use} of the extracted knowledge.

Therefore, in relations to our research question \textbf{RQ2}, we cannot conclude that the  procedural knowledge extracted from text is perceived as highly useful by our evaluators.

\textbf{Potential bias about AI.} As explained in Section~\ref{sec:setting}, we performed an A/B testing to check for any difference in people judgment, if they are told that the extraction task was executed by a LLM rather than an expert human annotator; in other words, we tested for any indication of a systematic bias about humans vs. machines.
For each of the evaluation items, we applied the Kruskal‐Wallis test~\cite{mckight2010kruskal} on the two groups (LLM vs. human annotator) to check for any difference. We run the test on the entire dataset (n=90 for each group), and the data displays only a few statistically significant differences between the two groups, in relation to items Q2, Q3, CQ1, CQ4 and U4 (p-values of the Kruskal-Wallis test between 0.007 and 0.04); in all those cases, the evaluators gave lower ratings to the LLM with respect to the expert human annotator. This result, even if not emerging on all items, seems to confirm the considerations out of our preliminary study.

In relations to our research question \textbf{RQ3}, therefore we cannot conclude that there is any systematic bias about LLMs, but a slight tendency of human evaluators to be more forgiving towards people than towards machines.

\textbf{Qualitative comparison between human and LLM knowledge extraction.} From a qualitative comparison between the annotations produced by human annotators and the knowledge extracted by the LLM, it emerges that both humans and LLMs benefit from the possibility to rephrase/simplify the original text of the procedure in order to generate instructions that are more concise and \emph{useful} from their point of view, even if this leads to a greater variability of the output. However, even when differently rephrased, the vast majority of annotators detected the same steps as the LLM.
Humans tended to rephrase also some verbs (e.g. \emph{turn on} in place of \emph{preheat}, \emph{prune} instead of \emph{nip off}) or some objects (e.g. \emph{packaging} instead of \emph{container}).
Furthermore, we noticed that both humans and the LLM took advantage of the option to include in their annotations implicit equipment items (e.g. inferring \emph{screwdriver} to check if screws are tight) -- increasing variability again. However, the LLM is more compliant to the given instructions, since it only extracted implicit equipment, while some human annotators introduced also implicit time indications when they consider it relevant to execute the procedure.

In relations to our research question \textbf{RQ4}, we can conclude that the extraction of both explicit and implicit knowledge from procedural texts shows multiple similarities between the LLM and the human annotators, with the only difference that humans are less likely to strictly follow the given instructions.

%
\section{Conclusions}
\label{sec:concl}


In this paper, we provided our results in the use of LLMs to extract procedural knowledge from textual descriptions, in order to build procedural knowledge graphs, adopting a suitable prompt chaining approach. 
Moreover, we performed an extensive human evaluation study to assess the LLM output in terms of its perceived quality and usefulness. We showed that the evaluators rated quite positively the quality of the extracted PK, while they were more doubtful about its usefulness; moreover, we highlighted that in some cases, people tend to be more critical towards an AI system rather than another human annotator: even if we did not find evidence of a systematic bias, this phenomenon is worth exploring in future LLM evaluation studies. Finally, we qualitatively studied the ability of both humans and LLMs to extract explicit and implicit knowledge from text, coming to the conclusion that, in extraction tasks where a definitive ground truth does not exist, an LLM displays abilities similar to those of people. 

All in all, we believe that LLMs are a promising technology to address this complex task, still the human intervention -- through a human-in-the-loop approach -- is very likely required to verify that the generated output is \enquote{good enough} to be used in real settings, especially when compliance is critical, like for example in the support to execute industrial procedures.

Our future work will be oriented towards a broader and more systematic evaluation of the proposed prompting approach, extending the assessment to industrial procedures in multiple formats (e.g. PDF or spreadsheets documents), addressing the challenge of extracting more complex cases of procedures (e.g. sub-procedures, optional or alternative steps, identification of additional entities), extending the approach to include fine-tuning or retrieval-augmented generation with the support of background knowledge, and comparing and benchmarking the outcomes of different LLMs.

\small \subsubsection*{Acknowledgements.} This work is partially supported by the PERKS project, co-funded by the European Commission (Grant id 101120323). The authors would like to thank all the participants to the human assessment reported in this paper.


%
%
%
\bibliographystyle{splncs04}
\bibliography{references.bib}
%


\end{document}